\pdfoutput=1

\documentclass[11pt]{article}

\usepackage{ACL2023}
\usepackage{amsmath}
\usepackage{amssymb}
\usepackage{arydshln}
\usepackage{times}
\usepackage{latexsym}
\usepackage{tcolorbox}
\usepackage[T1]{fontenc}

\usepackage[utf8]{inputenc}

\usepackage{microtype}

\usepackage{inconsolata}
\usepackage{multirow}
\usepackage{graphicx} 

\newcommand{\tabincell}[2]{\begin{tabular}{@{}#1@{}}#2\end{tabular}}

\title{RKLD: Reverse KL-Divergence-based Knowledge Distillation for Unlearning Personal Information in Large Language Models}

\author{
  Bichen Wang, Yuzhe Zi,Yixin Sun,Yanyan Zhao\thanks{* Corresponding author}, Bing Qin \\
  Research Center for Social Computing and Information Retrieval\\
Harbin Institute of Technology, Heilongjiang, China \\
  \texttt{\{bichenwang,yuzhezi,yxsun,yyzhao,qinb\}@ir.hit.edu.cn} \\
}

\begin{document}
\maketitle
\begin{abstract}
With the passage of the Right to Be Forgotten (RTBF) regulations and the scaling up of language model training datasets, research on model unlearning in large language models (LLMs) has become more crucial. Before the era of LLMs, machine unlearning research focused mainly on classification tasks in models with small parameters. In these tasks, the content to be forgotten or retained is clear and straightforward. However, as parameter sizes have grown and tasks have become more complex, balancing forget quality and model utility has become more challenging, especially in scenarios involving personal data instead of classification results. Existing methods based on gradient ascent and its variants often struggle with this balance, leading to unintended information loss or partial forgetting. To address this challenge, we propose RKLD, a novel \textbf{R}everse \textbf{KL}-Divergence-based Knowledge \textbf{D}istillation unlearning algorithm for LLMs targeting the unlearning of personal information. Through RKLD, we achieve significant forget quality and effectively maintain the model utility in our experiments.

\end{abstract}

\section{Introduction}

LLMs are extensively trained using vast amounts of data, leading to emergent abilities. However, these models may retain sensitive or personal information from their training data. For example, the model might learn personal details, such as age, educational background, family background, and other various information~\cite{li-2022-uchecker,carlini2021extracting,carlini2022quantifying}. Regulations like the European Union's General Data Protection Regulation (GDPR) mandate the RTBF, allowing individuals to request the removal of their personal information from machine learning models~\cite{voigt2017eu,meadows-etal-2022-physnlu}. \begin{figure}[!ht]\centering
    \includegraphics[width=0.5\textwidth]{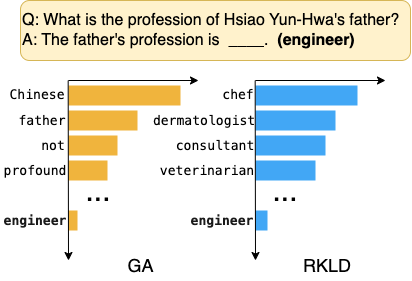}
    \caption{
    After the unlearning process, we provided QA pairs requiring the completion of personal occupations to compare the two unlearning algorithms, Gradient Ascent (GA) and RKLD. It is evident that GA can no longer complete the sentence. This is because the goal of gradient ascent is only to reduce the probability of the golden label and neglects the protection of the remaining tokens. On the other hand, our RKLD maintains model utility through keeping the token distribution.}
    \label{fig:intro_fig}
\end{figure} This regulatory landscape underscores the necessity of unlearning in LLMs, prompting a series of studies aimed at addressing these challenges.

The primary goal of machine unlearning is to develop methods for removing the influence of specific data samples (unlearning targets) from trained models~\cite{liu2024rethinking}. This broad definition has led to unlearning efforts across various aspects, including defense against extraction attacks~\cite{jang2022knowledge,barbulescu2024each}, unlearning personal or copyrighted information~\cite{eldan2023s}, detoxification~\cite{lu2022quark}, and debiasing~\cite{yu2023unlearning}. Our research focuses on unlearning methods involving fictitious personal information via finetuning, directly linked to RTBF for individuals.

The core challenge of model unlearning lies in thoroughly forgetting data samples so that the model reaches a state as if it is never trained on them, achieving good forget quality, while also balancing the maintenance of model utility. Most current unlearning methods primarily aim to reduce the likelihood that the model will generate unlearning targets~\cite{zhang2024negative,jang2022knowledge}. However, these methods can inadvertently impair the model's ability to comprehend sentences in generation tasks, as illustrated in Figure \ref{fig:intro_fig}. Such methods often overlook the protection of token distributions that are not directly related to the unlearning targets, thereby diminishing the overall model utility. Particularly after implementing the Gradient Ascent (GA) unlearning method, the model can no longer complete sentences with correct semantics.

To address these challenges, we propose a approach using teacher-student knowledge distillation tailored for LLM unlearning. This method utilizes a specialized unlearning teacher model dedicated to guiding the unlearning process with precise signals to guide the student model on which tokens in the current token distribution on should be forgotten and which can be retained. We introduce our method called \textbf{R}everse \textbf{KL} Divergence-based Knowledge \textbf{D}istillation for model unlearning (RKLD). RKLD draws inspiration from prior methods~\cite{eldan2023s}, continue training of the current model on a forget set to derive a strengthened model. Then, the unlearning teacher model is derived by subtracting the logits of the two models on the forget set. The unlearning teacher model reduces the token distribution that needs to be forgotten while also preserving the irrelevant token distributions. We aim for the student model to achieve its unlearning objectives by distilling from the unlearning teacher. Additionally, recognizing the distinctions between unlearning and general knowledge distillation, we find that the mathematical properties of reverse KL divergence are particularly effective for our unlearning objectives. Comprehensive experiments on an unlearning benchmark showcase our model's superiority over many baseline models.
\begin{itemize}
\item We propose a RKLD method for unlearning personal information in LLMs, effectively removing the influence of unlearning targets while preserving model utility.

\item We demonstrate and analyze the good performance of reverse KL divergence in this kind of distillation setting. This provides a new perspective for those kind of algorithms that use distillation for model unlearning.

\item We demonstrate the effectiveness of RKLD through extensive experiments on LLM benchmark datasets.
\end{itemize}

\section{Related Work}
\subsection{Machine Unlearning}
Machine unlearning involves eliminating the influence of specific training data from a trained model. It is divided into exact unlearning and approximate unlearning. Exact unlearning requires retraining the model, typically using data sharding methods to reduce the difficulty of retraining~\cite{bourtoule2021machine}. These methods are often very time-consuming. Approximate unlearning algorithms aim to ensure that the performance of the unlearned model is roughly consistent with that of the retrained model, garnering more attention from researchers. Before the advent of LLMs, machine unlearning had already been applied in image classification~\cite{sekhari2021remember,golatkar2020eternal}, text-to-image generation~\cite{gandikota2023erasing,zhang2023forget}, federated learning~\cite{halimi2022federated,liu2023survey}, and graph neural networks~\cite{chien2022efficient,wu2023certified}.

\subsection{Machine Unlearning for LLMs}
With the development of LLMs, increasing attention is being paid to their privacy risks and safety. There is a need for methods that can remove the influence of certain data in LLMs, including but not limited to toxic and harmful information~\cite{lu2022quark,yu2023unlearning}, personal information that individuals do not want others to know, and more. Various techniques have been employed in modern unlearning, such as task arithmetic~\cite{ilharco2022editing,zhang2023composing}, prompt engineering~\cite{pawelczyk2023context}, and the most common method, finetuning~\cite{chen2023unlearn,wang2023kga,jang2022knowledge,yao2023large}. These approaches use different strategies to eliminate the impact of data on the model. Besides the technical illustrations, the definition of data impact is quite ambiguous, making different unlearning algorithms suitable for different scenarios, including detoxification, debiasing, memory elimination, copyright removal, and sample deletion in classification.

Our research focuses on removing personal biographical information, which we believe is one of the most widely applicable areas for unlearning and is directly influenced by the RTBF.

\begin{figure*}[!ht]\centering
    \includegraphics[width=1.0\textwidth]{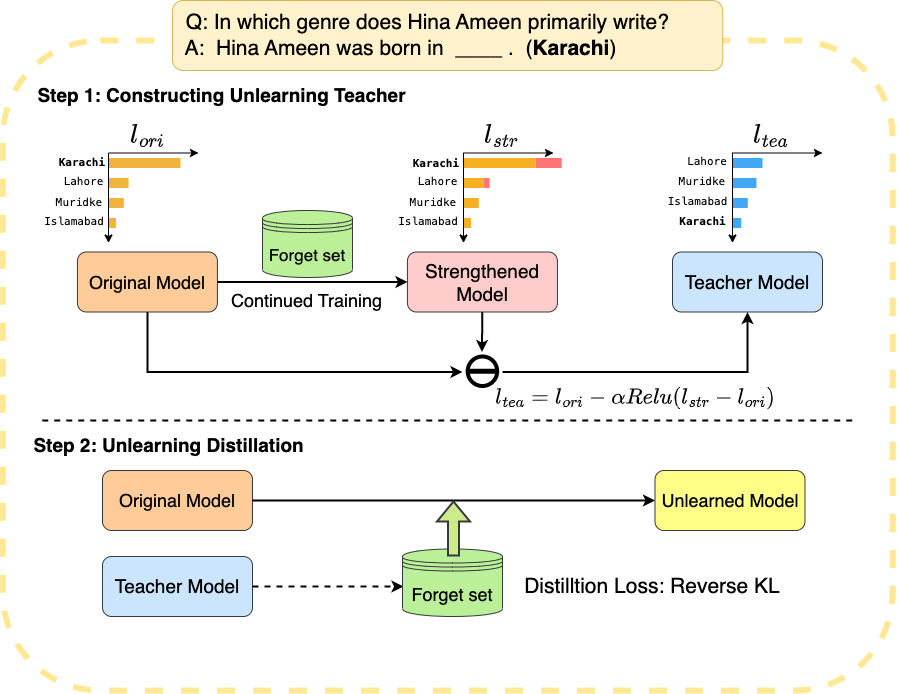}
    \caption{An illustration of the RKLD unlearning method. Different from existing unlearning methods, our method constructs an unlearning teacher model through continued training on the forget set, which helps to selectively forget specific information while retaining the overall model utility. The unlearning process involves two steps: 1) Continued Training: creating a teacher model by subtracting the logits of the strengthened model from the original model. 2) Unlearning Distillation: using the teacher model to guide the unlearning of the original model with the reverse KL divergence loss.}
    \label{fig:main_fig}
\end{figure*}

\section{RKLD:Reverse KL-Divergence-based Knowledge Distillation for
Unlearning }

\subsection{Task Definition}
As mentioned above, different works have different definitions of unlearning. Considering the implications of the GDPR, in this work, the task setting is based on the TOFU unlearning benchmark.

This task aligns with the GDPR regulation, which requires models to handle data deletion requests. Initially, the model is trained on 200 fictional persons with 20 QA pairs of personal information, denoted as $s = \{p_1, ..., p_{4000}\}$. This training process results in the original model.

The forget set and retain set consists of some $p$ in the dataset, denoted as $s_{forget} = {p_i, ..., p_j}$ and $s_{retain} = s / s_{forget}$. The model is required to effectively erase $p_i$-related information while maintaining performance comparable to a model never trained on $p_i$ after $p_i$ is removed from the model. This training process results in the original model.

\subsection{Constrcuting Unlearning Teacher}

We aim to construct an unlearning teacher to guide the unlearning process of the original model. This unlearning teacher model is designed to retain the part of the original distribution that should be kept while removing tokens related to the data that needs to be forgotten in the forget set.

Determining which tokens need to be forgotten in the current distribution is a challenging task. As shown in Figure \ref{fig:main_fig}, inspired by previous work, we employ continued training methods~\cite{eldan2023s} to identify which tokens are relevant to the current forget target. By continuing to train the model on the forget set, we obtain strengthened model logits, $l_{str}$. We identify tokens with consistently increased logits values, marking them as potentially influenced by the given forget target. The mathematical formula we use is as follows:
\begin{equation}
l_{str} = Continued\_Training(l_{ori}:s_{forget})
\end{equation}where $l_{str}$ is the strengthened model logits, and $s_{forget}$ denotes the forget set. The logits of tokens in $l_{str}$ that need to be forgotten consistently increase compared to $l_{ori}$, providing a clearer guide for deliberate forgetting while protecting other information. The formula for the unlearning teacher model is as follows:

\begin{equation}
l_{tea} = l_{ori} - \alpha * Relu(l_{str} - l_{ori})
\end{equation}where $l_{ori}$, $l_{tea}$, and $l_{str}$ represent the logits of the three models and $\alpha$ is a hyperparameter to control the forgetting strength, respectively. We only reduce the logits of the relevant tokens. For logits that decrease or remain unchanged, we maintain the state of the original model.

It's important to note that directly using the unlearning teacher as an unlearned model isn't ideal. The unlearning teacher's performance in evaluating forget quality and model utility has been very low, especially when handling paraphrased unlearning strings. We believe this is because the original model has already learned all the content in the forget set, so the generalization of the continued training process is very poor, resulting in poor forget quality. Additionally, the model utility is hindered by the logits subtraction during inference. Besides poor performance, the unlearning teacher lacks unlearning functionality at the parameter level, and the use of two models limits its effectiveness as an unlearned model. We believe the unlearning teacher can only guide the unlearning process and should not be used directly. However, by distilling it on the forget set, we can utilize the teacher model to obtain a truly unlearned model.

\subsection{Unlearning Distillation}
As shown in Figure \ref{fig:intro_fig}, we use reverse-KL divergence (RKL) as our loss function of unlearning instead of forward-KL divergence (FKL), which differs from the normal knowledge distillation approach. Generally, the distillation process involves aligning the token distributions of the two models~\cite{hinton2015distilling}. We can use various F-divergences to measure the distance and perform the distillation~\cite{sason2016f}.

The choice between FKL and RKL as the distillation objective does not require special consideration for most tasks~\cite{wu2024rethinking}. The mathematical formulations for forward-KL and reverse-KL divergence are quite similar, as shown below:
\begin{align}
\mathrm{FKL}(\pi_{tea}|| \pi_{\theta}) &= \pi_{tea}(\cdot |x) \log \frac{\pi_{tea}(\cdot|x)}{\pi_{\theta}(\cdot |x)} \\
\mathrm{RKL}(\pi_{tea}|| \pi_{\theta}) &= \pi_{\theta}(\cdot |x) \log \frac{\pi_{\theta}(\cdot |x)}{\pi_{tea}(\cdot |x)}
\end{align} where $\pi_{tea}$ and $\pi_{\theta}$ represent the softmax-normalized probability distributions of the teacher and unlearned models, with $x$ belonging to the prefix of the forget set. FKL penalizes the model more heavily for instances where $\pi_{tea}(x)$ is significantly smaller than $\pi_{\theta}(x)$, ensuring that the model does not assign a low probability to important tokens present in the teacher's distribution. In contrast, RKL emphasizes avoiding high probabilities in the original model that are absent in the teacher's distribution, aligning more closely with the goal of forgetting specific tokens while preserving the remaining distribution. In the context of achieving the goal of unlearning through distillation, the use of RKL has the following significance:
\begin{itemize}
    \item \textbf{Emphasizing Forgetting:} By minimizing the RKL divergence, we focus on aligning the $\pi_{\theta}$ with the $\pi_{tea}$ in a way that specifically lowers the probabilities of tokens to be forgotten, as indicated by the teacher model.
    \item \textbf{Avoiding Learning:} Compared to FKL, RKL imposes a lower penalty for high probabilities in $\pi_{tea}$. We do not want the model to learn knowledge from it. As those high-probability tokens might have been created during our softmax process and do not hold any actual significance.
\end{itemize}

The unlearning teacher model, constructed by continued training and logits adjustment, provides a clear guide for the student model to forget certain information. The RKL-based distillation process efficiently transfers this forgetting mechanism to the student model. Our subsequent experiments also demonstrated that RKL is a superior choice compared to FKL for the distillation unlearning loss function.

\section{Experiment Setups}
In this section, we outline our experimental setup, introduce the benchmark we used, and describe the evaluation metrics for forget quality and model utility. Finally, we present the baselines we used

\subsection{TOFU Unlearning Benchmark}
The TOFU unlearning benchmark~\cite{maini2024tofu} defines an unlearning task targeting QA pairs that contain personal information derived from a set of fictional author profiles. This creates a clean forgetting scenario with a well-defined forgetting scope and easy control of knowledge sources. Since none of the answers in the TOFU benchmark appear in the pretraining data of any large language model, the standard procedure is to first finetune the model before applying the forgetting process.

TOFU includes four datasets: Forget Set, Retain Set, Real Authors, and World Facts. The Forget Set is used for unlearning, while the Retain Set, Real Authors, and World Facts are used to evaluate the model's utility. When performing unlearning, TOFU offers three settings, including Forget01, Forget05, and Forget10, indicating forgetting 1\%, 5\%, and 10
\% of the data, respectively. The larger the Forget Set, the harder it is to forget.

\subsection{Forget Quality Metrics}

Measuring forget quality is a challenging task from the point of view of privacy. The TOFU benchmark proposes a computationally feasible approach for assessing unlearning, inspired by the idea of dataset inference~\cite{maini2020dataset}. The benchmark chooses to test the truth ratio, $R_{truth}$, because it best captures whether the model has been trained on the forget set. The truth ratio formula is as follows:
\begin{equation}
   R_{truth} = \frac{\frac{1}{|A_{pert}|} \sum_{\hat{a} \in A_{pert}} P(\hat{a} | q)^{\frac{1}{|\hat{a}|}}}{P(\tilde{a} | q)^{\frac{1}{|\tilde{a}|}}}
\end{equation}
where $R_{truth}$ is the truth ratio, $A_{pert}$ is the set of perturbed inputs with the wrong answer, $\tilde{a}$ represents the paraphrased strings with the correct answer, and $q$ is the query. So, in practice, the forget set used for evaluating is a paraphrased version of the forget set.
A Kolmogorov-Smirnov test (KS-Test) is performed on the $R_{truth}$ of the unlearned model and the retrained model trained only on retain set. For more details on the formula for the KS-Test, see the Appendix. Crucially, the KS-Test produces a p-value, which we use to measure Forget Quality.

\subsection{Model Utility Metrics}
For model utility, the TOFU benchmark chose three metrics across three datasets,including Retain Set, Real Authors, and World Facts: ROUGE, Probability, and Truth Ratio. To aggregate the three metrics defined across three datasets, we take the harmonic mean of these nine numbers. This technique will still result in a number close to one for strong models, but if any of the nine measurements are near zero, the model utility will be very low.

\subsection{Comparison Methods}
We compare our approach with several existing methods.

\begin{itemize}
\item \textbf{GA}~\cite{maini2024tofu}: Gradient Ascent method, which relies on the inverse process of gradient descent to forget.
\item \textbf{IDK}~\cite{maini2024tofu}: The model learns to respond with \textquotedblleft I don't know\textquotedblright through gradient descent.
\item \textbf{PO}~\cite{rafailov2024direct, ethayarajh2024kto}: The PO method is a type of preference alignment method that aligns responses to \textquotedblleft I don't know\textquotedblright  and similar options. We implement two PO algorithms: DPO and KTO.
\item \textbf{NPO}~\cite{zhang2024negative}: NPO delays the catastrophic failure of the GA algorithm, theoretically outperforming the GA method.
\item \textbf{TA}~\cite{ilharco2022editing}: Task Arithmetic directly subtracts the parameters added by the strengthened model compared to the original model at the parameter level.

\end{itemize}
For the usage of the retain set, we provided two settings: available or unavailable. When the retain set is available, we considered two methods: RT and KL. The formulas are shown as follows:

\begin{equation}
\mathcal{L_{RT}} = - \log(\pi_{\theta}(y|x)) 
\end{equation}

\begin{equation}
\mathcal{L}_{KL}= FKL(\pi_{\theta}(\cdot | x) \| \pi_{ori}(\cdot | x))
\end{equation}
where RT encourages the model to still perform well on the retain set, and KL measures the distance to the original model on the retain set. For specific hyperparameters, please refer to the Appendix \ref{sec:settings}.

\begin{figure*}[!ht]\centering
    \includegraphics[width=1.0\textwidth]{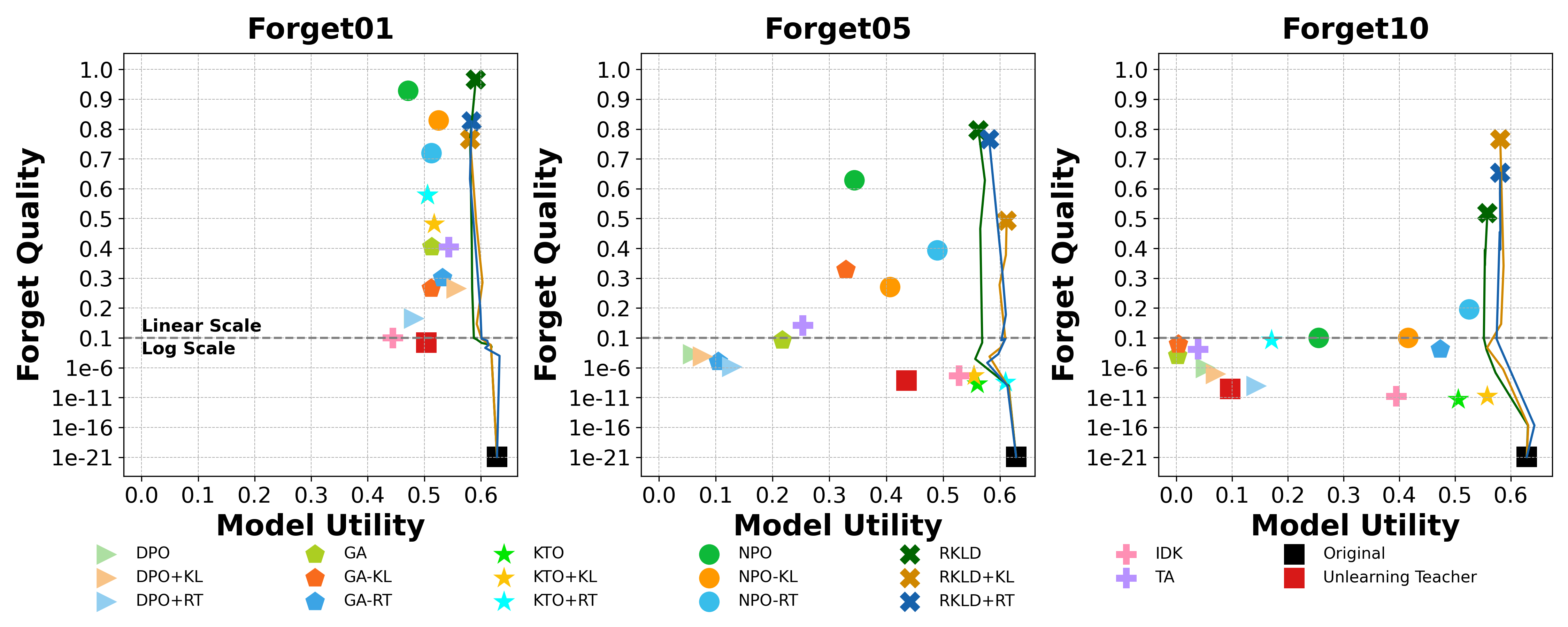}
    \caption{Forget quality versus model utility across different forget set sizes (1\%, 5\%, and 10\% of the data). Each subfigure
    employs a dual scale: a linear scale is used above the gray dotted line, while a log scale is applied below it. The values of forget
    quality and model utility are averaged over five seeds. Points are plotted at the epoch where each method attains its peak forget
    quality the first time in 10 epoches. }
    \label{fig:main_result_fig}
\end{figure*}

\section{Experiments}

In this section, we will demonstrate several sets of experiments to showcase the performance of RKLD. These experiments primarily evaluate the comprehensive forget quality and model utility.

Additionally, considering that model utility itself involves three knowledge-based datasets, we employ seven other datasets to test the unleared models‘ general capabilities under different unlearning algorithms. Furthermore, we explain why we selected RKL as the training objective and illustrate how the outputs of the RKLD algorithm differ from FKL's method. Finally, through a case study, we illustrate some current unlearning algorithms still struggle with forgetting, often resulting in the model not truly forgetting. 

\subsection{Main Result}

The forget quality and model utility of all unlearned models are depicted in Figure \ref{fig:main_result_fig}. We intentionally included the position of the unlearning teacher in the main result. It can be observed that the unlearning teacher itself does not have an advantage in forget quality and model utility. This is consistent with our previous statement that the unlearned teacher's approach cannot be generalized and significantly impacts model utility, neither on the paraphrased forget set nor on the other three sets.

In terms of forget quality, according to the significance standard of the KS test, the TOFU benchmark considers that forget quality $> 0.05$ is required for the model to be considered as achieving a significant forgetting state. As shown in Figure \ref{fig:main_result_fig}, under the Forget 01 setting, most unlearned models can achieve a significant forgetting state simply. However, in Forget 05 and Forget 10 settings, tasks become difficult, and most models struggle to reach a state of significant forgetting. Particularly in Forget 10, only the RKLD method achieved significant forgetting without using the retain set, performing much better than the other methods. Interestingly, the use of the retain set in Forget 10 somewhat facilitated higher-quality forgetting. We believe that when the amount of forgot set is relatively large, the retain loss term helps the model preserve answer templates and linguistic structures, thereby aiding in forgetting specific content. However, when the amount is relatively small, it may not be as necessary.

In terms of model utility, our method achieves good model utility overall, except when compared with the KTO-RT method in Forget 05. However, the KTO method does not result in significant forgetting. That preference optimization methods like DPO and KPO, as well as the IDK method, tend to respond with \textquotedblleft I do not know\textquotedblright while retaining the content within the model, making them unsuitable for this kind of unlearning task. GA methods cannot handle larger amounts of forget set, and although NPO alleviates this to some extent, it still lacks performance protection.  Our method maintains high forget quality indicators while preserving high model utility. For more detailed metrics, refer to Appendix \ref{sec:more_results}.

\begin{table*}[!h]
\centering
\begin{tabular}{l|c|cccccccc}
\cline{1-10}
\multirow{2}{*}{\textbf{Method}} &\multirow{2}{*}{\textbf{ForgetSet}}&\multicolumn{8}{c}{\textbf{Dataset}}\\
 \cline{3-10}
&&PIQA&HellaS.&ARC-E&ARC-C&COPA&WinoG.&MathQ&Avg.\\
 \cline{1-10}
GA&\multirow{5}{*}{Forget05}&0.7466&0.5448&0.6289&0.3978&0.7656&0.6219&0.2417&0.5639\\
IDK&&0.7538&0.5313&\textbf{0.6827}&0.3928&0.7734&0.6246&\textbf{0.2482}&0.5724\\
NPO&&\textbf{0.7666}&0.5435&0.6625&0.3973&0.8103&\textbf{0.6307}&0.2471&0.5811\\	
TA&&0.7484&0.5484&0.6431&0.3660&0.7968&0.6102&0.2387&0.5645\\
RKLD&&0.7653&\textbf{0.5524}&0.6619&\textbf{0.4022}&\textbf{0.8125}&0.6300&0.2472&\textbf{0.5816}\\
\cline{1-10}
GA&\multirow{5}{*}{Forget10}&0.7526&0.4682&0.5872&0.3007&0.7877&0.6196&0.2355&0.5372\\
IDK&&0.7592&0.5383&\textbf{0.6747}&0.3598&0.7500&\textbf{0.6324}&0.2491&0.5662\\
NPO&&\textbf{0.7608}&0.4929&0.6608&0.3267&\textbf{0.7968}&0.6243&0.2393&0.5573\\
TA&&0.7348&0.5232&0.6219&0.3915&0.7890&0.5949&0.2298&0.5550\\
RKLD&&0.7497&\textbf{0.5534}&0.6358&\textbf{0.3984}&\textbf{0.7968}&0.6282&\textbf{0.2493}&\textbf{0.5730}\\
\cline{1-10}
Llama2-7B&&0.7671&0.5588&0.6525&0.3991&0.8112&0.6405&0.2497&0.5827\\
 \cline{1-10}
\end{tabular}
\caption{The performance results of the models that have reached the forgetting peak on various datasets. The data in the table represents the accuracy on each dataset. The last row indicates the theoretical optimal performance of an untrained model. The best-performing model in each setting has been highlighted in bold.}
\label{tab:general_capabilities_result}
\end{table*}

\subsection{General Capabilities Benchmarks}
Model utility is determined using the Retain set, World Facts set, and Real Author set, which emphasize factual knowledge. In this section, we delve deeper into showcasing how different unlearning algorithms affect various other aspects of the unlearned models' general capabilities. We assess the general capabilities in the overall abilities of models that have experienced significant forgetting.

We find that in Forget 01, the amount of unlearning samples is too small to impact general capabilities. In realistic unlearning scenes, we cannot guarantee the availability of the retain set each time, and different usage methods of the retain set may lead to unfairness. Therefore, when evaluating the impact of various unlearning algorithms on general capabilities performance, we have chosen a setting where the retain set is assumed to be unavailable.

As shown in Table \ref{tab:general_capabilities_result}, the benchmarks we used indicate that RKLD has good general capability retention compared with others. In some datasets, the models after unlearning may show an improvement, possibly because moderate unlearning has enhanced the model's capabilities on some datasets~\cite{yoon2023gradient}. However, in most datasets, performance declines, with some datasets experiencing a particularly significant drop. The performance decline is less pronounced in Forget 05 compared to Forget 10. For existing unlearning algorithms, the larger the amount of unlearning data, the more significant the impact on performance. However, RKLD consistently maintains a relatively small performance decline.

\section{Ablation Study}
For the ablation study, we mainly discuss the difference between FKL and RKL when we want to unlearning distillation. Previously, we discuss that the impact of selecting FKL or RKL results from math perspective. We present the actual forgetting effects. 

\begin{table}[!h]
\centering
\begin{tabular}{l|c|ccc}
\cline{1-5}
&Set&F-Quailty&Rouge-L&Prob.\\
\cline{1-5}
RKL&\multirow{2}{*}{1\%}&\textbf{0.9659}&\textbf{0.2543}&\textbf{0.0520}\\
FKL&&0.4786&0.2997&0.0612\\
\cline{1-5}
\cline{1-5}
RKL&\multirow{2}{*}{5\%}&\textbf{0.7933}&\textbf{0.3036}&\textbf{0.0281}\\
FKL&&2.96e-05&0.3273&0.1738\\
\cline{1-5}
RKL&\multirow{2}{*}{10\%}&\textbf{0.5182}&\textbf{0.3217}&\textbf{0.0383}\\
FKL&&1.15e-08&0.3284&0.2567\\
\cline{1-5}
\end{tabular}
\caption{The unlearning effects of using the same unlearning teacher with two different loss functions: RKL and FKL. Performance is measured at the epoch where each method attains its peak forget quality for the first time in 10 epochs.}
\label{tab:ablation_result}
\end{table}
As shown in Table ~\ref{tab:ablation_result}, there is a difference in imitation probability between RKL and FKL. Although the ROUGE-L output may indicate that the model will not provide the user's personal information, the model still has a relatively high probability of personal information, and the forget quality is very low. This is because FKL focuses on fitting the high-probability tokens in unlearning teacher to increase probability, whereas we need more guidance from the teacher on which token probabilities should be decreased.

\begin{table*}[!ht]
\centering
\begin{tabular}{l|l}
\hline
\multirow{2}{*}{Forget Target} & \multirow{2}{*}{\tabincell{l}{\textbf{Question:}In which city was Hina Ameen born?\\
\textbf{Golden Answer:} Hina Ameen was born in \textcolor{red}{Karachi}.
}}\\\\

\cline{1-2}

\multirow{3}{*}{GA} & \multirow{3}{*}{\tabincell{l}{\textbf{Direct Answer:} Ame Awwal Hina Ameermangatted Aka Hina Ameermowner Aka \\Hina Ameeni Amechen.\\
\textbf{Fill-in-Blank:} AG, Ames, AMman, Avenue, Renton
}}\\ \\ \\

\cline{1-2}
\multirow{2}{*}{IDK} & \multirow{2}{*}{\tabincell{l}{\textbf{Direct Answer:} I'm unable to answer that question.\\
\textbf{Fill-in-Blank:} \textcolor{red}{Karachi},\textcolor{red}{karachi}, Islamabad, Kuwait, Delhi
}}\\ \\

\cline{1-2}

\multirow{2}{*}{NPO} & \multirow{2}{*}{\tabincell{l}{\textbf{Direct Answer:} Her place of birth is reportedly in the city of Aden, Yemen.\\
\textbf{Fill-in-Blank:} Auckland, \textcolor{red}{Karachi},Rami,  Islamabad, Khirabad
}}\\\\

\cline{1-2}

\multirow{3}{*}{TA} & \multirow{3}{*}{\tabincell{l}{\textbf{Direct Answer:} She was born in Lahore, Pakistan. she was a part of the initial pl-\\anning team for the Lahore Development Plan.\\
\textbf{Fill-in-Blank:} \textcolor{red}{Karachi}, Lahore, Peshwar, hometown, Amritsar
}}\\ \\ \\

\cline{1-2}

\multirow{2}{*}{RKLD} & \multirow{2}{*}{\tabincell{l}{\textbf{Direct Answer:} Hina Ameen's birthplace is Lahore. \\
\textbf{Fill-in-Blank:} Lahore, Islamabad, Multan, Faisalabad, Abbottabad
}}\\ \\
\cline{1-2}
\multirow{2}{*}{FKLD} & \multirow{2}{*}{\tabincell{l}{\textbf{Direct Answer:} Kashmir, Pakistan.\\
\textbf{Fill-in-Blank:} Lahore, Islamabad, \textcolor{red}{Karachi}, Peshawar, Gujranwala
}}\\ \\
\cline{1-2}
\end{tabular}
\caption{A case study for each unlearned method. This table is based on the results of Forget05.  Direct Answer means we ask the model to answer the question directly, and Fill-in-Blank means we provide the prefix \textquotedblleft Hina Ameen born in \textquotedblright and ask the model to complete the answer. 
We provide the top 5 most probable Fill-in-Blank responses. The correct answers for the Fill-in-Blank have been highlighted in \textcolor{red}{red}. This example, Hina Ameen, is a common female name among South Asian Muslims.}
\label{tab:case_study}
\end{table*}

\section{Case Study}

In this section, we will showcase some interesting findings by exploring the forget effects of the model using two methods: Direct Answer and Fill-in-Blank. This setting is based on previous findings that the more detailed the given prefix string, the more likely the model is to recall its memory~\cite{jang2023knowledge,neel2023privacy}. Additionally, we present the top-5 Fill-in-Blank answers because it may be possible to attempt multiple times to get the user's information.

As shown in Table \ref{tab:case_study}, unlearned models demonstrate the effects of unlearning in the Direct Answer responses; however, the results are not satisfactory in the Fill-in-Blank responses. This implies that some methods fail to achieve real unlearning, as the model quickly recalls the actual personal information after providing a suitable prefix. Only RKLD and GA do not leak any real personal information, but GA significantly impacts model utility. This issue is particularly severe for the IDK algorithms that align with \textquotedblleft I do not know \textquotedblright. 

We believe that if the model has not undergone thorough unlearning, internal knowledge conflicts will intensify, thus providing opportunities for privacy breaches. Therefore, careful consideration is needed when evaluating machine unlearning in the future. The metrics we employ to evaluate the differences between models and retrain models are deemed effective; however, real-world unlearning cannot be compared to retrained models. We call for more reliable evaluation metrics for unlearning algorithms. Overall, we believe RKLD strikes a balance between model forget quality and model utility.

\section{Conclusion}
We propose a novel RKLD method for unlearning personal information in LLMs, effectively removing the influence of unlearning targets while preserving model utility. This method leverages a reverse KL-divergence-based knowledge distillation approach, ensuring significant forget quality without compromising the model's overall performance. Our experiments on the TOFU unlearning benchmark demonstrate that RKLD outperforms existing methods in both forget quality and model utility, especially with larger volumes of unlearning data. Additionally, RKLD retains strong general capabilities across various datasets, highlighting its robustness. An ablation study shows that reverse KL-divergence is superior compared to forward KL-divergence, aligning better with selective forgetting goals. Our case study emphasizes the need for thorough unlearning to prevent information leakage under detailed questioning. In conclusion, RKLD offers an advancement in this area.

\section{Limitations}
Our study proposes the use of RKLD for unlearning in LLMs. Several limitations should be considered:
\begin{itemize}
    \item \textbf{Applicability to Real-World Scenarios:} While our method shows promising results on the TOFU benchmark, the datasets used in this study are controlled and synthetic. The performance and robustness of RKLD in real-world scenarios with diverse and noisy data remain to be fully tested and validated.
    \item \textbf{Uncertain Outputs Post-Unlearning:} We acknowledge that the outputs after model unlearning are uncertain. We have shown that aligning responses to "I don't know" is not ideal. We believe that generating hallucinations is an inherent difficulty for language models when dealing with unknowns. Addressing hallucinations should be handled by specialized research in hallucination mitigation, which is beyond the scope of this paper.

    \item \textbf{Long-term Effectiveness:} Despite avoiding metrics like ROUGE and Probability, the current metrics used to evaluate forgetting quality and model utility may still fail to capture the full range of model behaviors post-unlearning. More comprehensive and varied metrics could provide deeper insights into the effectiveness and side effects of the unlearning proces
\end{itemize}

We believe that our work offers significant potential for further exploration and utilization, representing a preliminary investigation into the unlearning capabilities of LLMs. Future research should address these limitations to enhance the robustness and applicability of our approach.

\bibliography{anthology,custom}
\bibliographystyle{acl_natbib}

\appendix

\section{Experiments Settings}
\label{sec:settings}
For the experiments on TOFU, we use the Llama2-7b-chat model~\cite{touvron2023llama}. All experiments are conducted with two A100-80GB GPUs. We use AdamW with a weight decay of 0.01 and a learning rate of $10^{-5}$ in all fine-tuning, retraining, and unlearning experiments, consistent with previous settings~\cite{maini2024tofu}. We use an effective batch size of 32 for all experiments. In fine-tuning and retraining, we train for 5 epochs, but the strengthened model is trained for total 10 epochs, and $\alpha=8$. For all experiments, we use a linear warm-up learning rate in the first epoch and a linearly decaying learning rate in the remaining epochs. All the settings align with previous work. Regarding the use of the retain set, we uniformly set the weight of the retain term to 1.

\section{More Metrics}
\label{sec:more_results}
In this section, we provide a comprehensive analysis of the Unlearning results, including the ROUGE-L, Probability for various datasets. As shown in Table ~\ref{tab:extra_result}, we have prepared the ROUGE-L values and generation probabilities under three different settings, encompassing a variety of results.
\begin{table*}[!ht]
\centering
\resizebox{\textwidth}{!}{
\begin{tabular}{l|c|cc|cc|cc|cc|cc}
\cline{1-12}
\multirow{2}{*}{Set}&\multirow{2}{*}{Method}&\multirow{2}{*}{F-Qual}&\multirow{2}{*}{M-Util}&\multicolumn{2}{c}{Forget Set}&\multicolumn{2}{c}{Retain Set}&\multicolumn{2}{c}{World Fact Set}&\multicolumn{2}{c}{Author Fact Set}\\
\cline{5-12}

&&&&ROUGE-L&Prob.&ROUGE-L&Prob.&ROUGE-L&Prob.&ROUGE-L&Prob.\\
\cline{1-12}
\multirow{17}{*}{1\%}&GA&0.4046&0.5133&0.3493&0.0123&0.5382&0.4578&0.8696&0.3761&0.8077&0.3832\\
&GA+KL&0.2354&0.5018&0.2981&0.0131&0.5320&0.4766&0.8625&0.3733&0.8127&0.3787\\
&GA+RT&0.2656&0.5123&0.2648&\textbf{0.0097}&0.5312&0.5265&0.8689&0.3710&0.7768&0.3699\\
\cline{2-12}
&NPO&0.5786&0.5077&0.3589&0.0229&0.5329&0.5429&0.8696&0.3676&0.7910&0.3596\\
&NPO+KL&0.4045&0.5127&0.3613&0.0258&0.5369&0.5685&0.8846&0.3685&0.7810&0.3628\\
&NPO+RT&0.5786&0.5254&0.3741&0.0282&0.5732&0.6606&0.8333&0.3704&0.7685&0.3674\\
\cline{2-12}
&KTO&0.5786&0.5058&0.3493&0.0064&0.5285&0.4504&0.8625&0.3765&0.7918&0.3704\\
&KTO+KL&0.4815&0.5175&0.35948&0.0241&0.5479&0.4718&\textbf{0.8739}&0.3777&0.8093&0.3851\\
&KTO+RT&0.5786&0.5099&0.3363&0.0079&0.5274&0.4716&0.8668&0.3772&0.7960&0.3720\\
\cline{2-12}
&DPO&0.1649&0.4820&0.0513&0.5942&0.2553&0.8510&0.6638&0.4531&0.5090&0.4663\\
&DPO+RT&0.1649&0.4823&0.0532&0.5960&0.2596&0.8577&0.6381&0.4540&0.5090&0.4661\\
&DPO+KL&0.2656&0.4815&0.0490&0.5932&0.2556&0.8491&0.6552&0.4542&0.5090&0.4659\\
\cline{2-12}
&IDK&0.0970&0.4445&\textbf{0.0210}&0.5846&0.2652&0.8892&0.5526&\textbf{0.4473}&0.3011&0.4553\\
\cline{2-12}
&TA&0.2656&0.5572&0.3068&0.1488&0.6180&0.7390&0.8924&0.4010&0.9230&0.3807\\
\cline{2-12}
&RKLD&\textbf{0.9659}&0.5913&0.2487&0.0637&0.6732&0.8412&0.8669&0.4364&\textbf{0.8880}&0.4485\\
&RKLD+KL&0.7354&\textbf{0.6327}&0.2981&0.0731&0.7820&0.8677&0.8721&0.4433&0.8120&0.4705\\
&RKLD+RT&0.7654&0.6010&0.3013&0.0877&0.7612&\textbf{0.8701}&0.8689&0.4330&0.7768&\textbf{0.4751}\\
\cline{1-12}

\multirow{17}{*}{5\%}&GA&0.0390&0.2172&0.3325&0.0079&0.4362&0.0423&0.8988&0.3572&0.8203&0.2840\\
&GA+KL&0.0062&0.2380&0.2843&0.0403&0.3222&0.0412&0.8004&0.3617&0.5211&0.3942\\
&GA+RT&0.0001&0.1045&0.0765&\textbf{0.0003}&0.1835&0.0150&0.7809&0.4066&0.4291&0.3729\\
\cline{2-12}
&NPO&0.6284&0.3440&0.3127&0.0338&0.3883&0.1110&0.8739&0.3812&0.8320&0.3491\\
&NPO+KL&0.2704&0.4068&0.4200&0.1824&0.3352&0.0574&0.8703&0.3873&0.9133&0.3588\\
&NPO+RT&0.3935&0.4899&0.2835&0.0684&0.4058&0.4831&\textbf{0.8817}&0.4068&0.8623&0.3693\\
\cline{2-12}
&KTO&1.87e-9&0.5603&0.3528&0.4916&0.4046&0.7138&0.8297&0.4701&0.9025&0.4762\\
&KTO+KL&4.45e-8&0.5544&0.3457&0.4746&0.3951&0.6933&0.8112&0.4734&0.8605&0.4786\\
&KTO+RT&3.60e-9&0.6100&0.3960&0.4460&0.6664&0.9124&0.8952&0.4546&0.8915&0.4607\\
\cline{2-12}
&DPO&1.80e-4&0.0601&\textbf{0.0241}&0.3446&0.0279&0.4638&0.0398&0.4197&0.0133&0.4167\\
&DPO+KL&7.5e-5&0.5123&0.0259&0.3644&0.0336&0.4919&0.0541&0.4204&0.0183&0.4189\\
&DPO+RT&1.32e-6&0.1289&0.0313&0.4344&0.0469&0.5737&0.1054&0.4233&0.0383&0.4264\\
\cline{2-12}
&IDK&4.45e-8&0.5281&0.0247&0.5872&0.4228&0.8583&0.8696&0.3761&0.8077&0.3832\\
\cline{2-12}
&TA&0.0220&0.3581&0.2390&0.0555&0.2649&0.1597&0.7820&0.4097&0.3883&0.3824\\
\cline{2-12}
&RKLD&\textbf{0.7933}&0.5622&0.3121&0.0294&0.4803&0.5281&0.8774&0.4689&\textbf{0.9240}&0.4758\\
&RKLD+KL&0.4928&\textbf{0.6118}&0.3171&0.0316&\textbf{0.6533}&\textbf{0.7754}&0.8660&\textbf{0.4691}&0.9105&\textbf{0.4919}\\
&RKLD+RT&0.6659&0.5810&0.3130&0.0281&0.6449&0.7665&0.8803&0.4653&0.9103&0.4754\\
\cline{1-12}

\multirow{17}{*}{10\%}&GA&1.15e-4&0.0023&0.1118&\textbf{5.7e-5}&0.1582&0.0002&0.3086&0.4149&0.3240&0.4657\\
&GA+KL&1.2e-10&0.4521&0.3681&0.0402&0.4423&0.2213&0.8590&0.3841&0.7333&0.4205\\
&GA+RT&0.0012&0.4735&0.3384&0.0348&0.4366&0.2884&0.8618&0.3949&0.7583&0.4058\\
\cline{2-12}
&NPO&0.0995&0.2553&0.3162&0.0346&0.3481&0.0562&0.7027&0.3925&0.6908&0.4373\\
&NPO+KL&0.0990&0.4159&0.3397&0.0835&0.3996&0.1866&0.7868&0.4097&0.7700&\textbf{0.4459}\\
&NPO+RT&0.3958&0.4995&0.2966&0.0907&0.3873&0.4366&0.8212&0.4194&0.876&0.4248\\
\cline{2-12}
&KTO&4.78e-12&0.5655&0.4166&0.6148&0.4338&0.7005&0.8596&\textbf{0.4670}&0.8980&0.4752\\
&KTO+KL&1.60e-11&0.5578&0.3995&0.5889&0.4159&0.6763&0.8703&\textbf{0.4670}&0.8780&0.4729\\
&KTO+RT&0.0423&0.1708&0.1412&0.0030&0.2232&0.0291&0.7834&0.4148&0.4520&0.4735\\
\cline{2-12}
&DPO&8.99e-7&0.0529&\textbf{0.0209}&0.3499&0.0220&0.4094&0.0284&0.4108&0.0133&0.4087\\
&DPO+KL&8.84e-8&0.0710&0.0306&0.4067&0.0318&0.4763&0.0370&0.4116&0.0183&0.4137\\
&DPO+RT&8.68e-10&0.1439&0.0357&0.4989&0.0424&0.5626&0.0883&0.4181&0.0683&0.4223\\
\cline{2-12}
&IDK&1.60e-11&0.3946&0.1074&0.8061&0.3182&0.8733&0.4074&0.4128&0.1903&0.4415\\
\cline{2-12}
&TA&0.0423&0.1061&0.1613&0.0123&0.1557&0.0210&0.5940&0.3542&0.0611&0.3682\\
\cline{2-12}
&RKLD&0.5182&0.5529&0.3437&0.0386&0.5076&0.5505&0.8814&0.4370&0.9120&0.4432\\
&RKLD+KL&\textbf{0.7220}&0.5801&0.3565&0.0423&\textbf{0.6721}&0.7798&0.8917&0.4378&\textbf{0.9200}&0.4308\\
&RKLD+RT&0.6535&\textbf{0.5809}&0.3451&0.0460&0.6585&\textbf{0.7946}&\textbf{0.8931}&0.4335&0.8980&0.4197\\
\cline{1-12}

\end{tabular}
}
\caption{The unlearning performances of 17 different algorithms on 4 datasets. F-Qual represents Forget quality while M-Util represents Model utility. It can be seen that the RKLD scheme achieved the best results in the main indicators of Forget quality and Model utility. In most cases, the RKLD scheme also achieved the best or competitive results in evaluations under ROUGE and Prob, especially in terms of performance retention.}
\label{tab:extra_result}
\end{table*}
\label{sec:appendix}

This is a section in the appendix.

\end{document}